\documentclass[letterpaper]{article} 
\usepackage{aaai24}  
\usepackage{times}  
\usepackage{helvet}  
\usepackage{courier}  
\usepackage[hyphens]{url}  
\usepackage{graphicx} 
\def\UrlFont{\rm}  
\usepackage{natbib}  
\usepackage{caption} 
\usepackage{algorithm}
\usepackage{algorithmic}

\usepackage{newfloat}
\usepackage{listings}
\DeclareCaptionStyle{ruled}{labelfont=normalfont,labelsep=colon,strut=off} 
\floatstyle{ruled}
\newfloat{listing}{tb}{lst}{}
\floatname{listing}{Listing}
\usepackage{booktabs}       
\usepackage{amsmath}
\usepackage{amssymb}
\usepackage{mathtools}
\usepackage{amsthm}
\usepackage{algorithm}
\usepackage{algorithmic}
\usepackage{mathabx}
\usepackage{array}
\usepackage{enumitem}
\usepackage{multirow}
\usepackage{subfigure}
\setlist[enumerate]{nosep}
\usepackage{soul}
\usepackage{caption}
\usepackage{xcolor}

\theoremstyle{plain}
\newtheorem{theorem}{Theorem}[section]

\theoremstyle{definition}
\newtheorem{definition}[theorem]{Definition}

\theoremstyle{remark}

\newcommand{\x}{{\bf x}}
\newcommand{\w}{{\bf w}}

\newcommand{\STATEnonum}{\item[]}

\title{DeepSaDe: Learning Neural Networks that Guarantee Domain Constraint Satisfaction}
\author{
    Kshitij~Goyal\textsuperscript{\rm 1}, 
    Sebastijan Dumancic\textsuperscript{\rm 2},
    Hendrik~Blockeel\textsuperscript{\rm 1}
}
\affiliations{
    \textsuperscript{\rm 1}KU Leuven, Belgium\\
    \textsuperscript{\rm 2}Delft University of Technology, The Netherlands\\

    kgoyal40@gmail.com, s.dumancic@tudelft.nl, hendrik.blockeel@kuleuven.be
}

\usepackage{bibentry}

\begin{document}

\maketitle

\begin{abstract}
As machine learning models, specifically neural networks, are becoming increasingly popular, 
there are concerns regarding their trustworthiness, especially in safety-critical applications, e.g., actions of an autonomous vehicle must be {\it safe}.
There are approaches that can train neural networks where such domain requirements are enforced as constraints, but they either cannot guarantee that the constraint will be satisfied by all possible predictions (even on unseen data) or they are limited in the type of constraints that can be enforced.
In this work, we present an approach to train neural networks which can enforce a wide variety of constraints and guarantee that the constraint is satisfied by all possible predictions.
The approach builds on earlier work where learning linear models is formulated as a constraint satisfaction problem (CSP). To make this idea applicable to neural networks, two crucial new elements are added: constraint propagation over the network layers, and weight updates based on a mix of gradient descent and CSP solving.
Evaluation on various machine learning tasks demonstrates that our approach is flexible enough to enforce a wide variety of domain constraints and is able to guarantee them in neural networks.
\end{abstract}

\section{Introduction}
Widespread use of state-of-the-art machine learning (ML) techniques has given rise to concerns regarding the trustworthiness of these models, especially in safety-critical and socially-sensitive domains. For example, in autonomous vehicles that employ ML approaches to predict the next action, the actions must be safe. 
Such domain requirements can often be formulated as logical constraints on combinations of inputs and outputs (e.g., whenever the input satisfies some condition A, the output must satisfy condition B). Crucially, these domain constraints must be satisfied for all possible inputs to the model, not just the training data.  
This has motivated researchers to develop approaches that can train models that satisfy a given constraint for all possible predictions.
 
A general approach to enforcing constraints in ML models is to include a regularization term in the cost function, which typically adds a cost for every violation of a constraint in the training set (e.g., \citet{xu2018semantic,diligenti2017semantic}). 
Such an approach can reduce the number of violations in the training set, but it does not necessarily eliminate them. Moreover, even when it does, this does not guarantee that other instances (outside the training set) cannot violate the constraint.    
Alternatively, for some model types, such as neural networks, the architecture of the model can be chosen in such a way that certain types of constraints are guaranteed to be satisfied for each possible input (not just training data) \cite{sivaraman2020counterexample,hoernle2022multiplexnet}. But this is typically possible only for specific combinations of model and constraint types.

This raises the question of whether generality and certainty can be combined. Is it possible to come up with a {\em generally applicable} approach that guarantees the satisfaction of constraints not only on the training set but on the {\em entire} input space, and this for any kind of model?
A step in this direction was made by \citet{sade}, who propose a relatively general solution for linear models.  Their approach translates the learning problem into a MaxSMT setting.  MaxSMT stands for Maximum Satisfiability Modulo Theories. It is an extension of SAT solving that can take background theories into account (e.g., for reasoning about the real numbers) and that distinguishes soft and hard constraints: it returns a solution that satisfies all hard constraints and as many soft constraints as possible. 
\citet{sade} model the requirements as hard constraints and maximize the fit to the data using soft constraints.  
Their approach works for a wide range of constraint types but only handles linear models, and assumes a bounded input domain.
 
In this paper, we substantially extend the applicability of that approach by showing how it can be used to train feedforward neural networks. 
Two key modifications to the network’s architecture and training procedure suffice for achieving this: (1) propagating the constraints over the network layers to the last layer, which involves adding skip connections \cite{he2016deep} that copy the input to the penultimate layer and deriving bounds on the penultimate layer from the bounds on the input \cite{sunaga1958theory}, and (2) training the network using a hybrid procedure that uses MaxSMT to determine the weights in the output layer and gradient descent for all other weights. We demonstrate that with these changes, neural networks can be trained that have good performance and guarantee the satisfaction of given constraints.  
In the following, we first describe the problem setting, then briefly describe the existing approach that we build on, before detailing our approach. 
Afterward, we compare our approach to related work and evaluate it experimentally.

\section{Problem Statement}\label{rep:constraints}
In this paper, we focus on semantic constraints which constrain the behavior of the model: 
the predictions are required to adhere to certain requirements, 
e.g. safety constraints \cite{katz2017reluplex}, structured output constraints \cite{xu2018semantic}, and hierarchical constraints \cite{hoernle2022multiplexnet}.
Specifically, we focus on domain constraints: constraints that must hold for all instances in the domain. 
We assume the constraints can be written with universally quantified logic formulas.
In particular, we consider the constraints of form:
\begin{equation}
\text{K}: \forall \x \in \bigtimes_{i=1}^{n} [l_i, u_i], \x \models P \Rightarrow f_{\w}(\x) \models C 
\label{equation:constraint}
\end{equation}
Which states that if an input $\x$ satisfies a condition $P$, the output must satisfy a condition $C$.
An example of a safety constraint that can be represented in this way is: ``if an object comes in front of a moving vehicle, the vehicle must stop''.
Solving for such a constraint exactly using specific constraint solvers allows us to enforce the constraint for all possible inputs.
The reason for focusing on this type of constraint is mostly practical: the constraint-solving technology we use was found to scale well enough for these types of constraints.  
It is not a theoretical limitation: any constraint that can be handled effectively by current constraint-solving technology can be handled by the approach we develop.
As demonstrated later in section \ref{sec:experiment}, our chosen constraint formulation (equation \ref{equation:constraint}) already provides us with a variety of tasks to work with. 

To make the search procedure tractable, and because features in ML problems are typically bounded (e.g., a pixel in an image takes a value in $[0, 255]$), we use bounded domains $\bigtimes_{i=1}^{n}[l_i, u_i]$ instead of $\mathbb{R}^n$. As a natural choice, the training data $D \subseteq \mathcal{X} \bigtimes \mathcal{Y}$ can be used to calculate these bounds, i.e. $l_i = \min_{D}(\x_i)$ \& $u_i = \max_{D}(\x_i)$.
We rely on the SMT solver Z3 \cite{moura2008z3} to solve such constraints.
Other prominent constraint-solving paradigms like MILP and CP do not support such constraints over continuous domains \cite{nethercote2007minizinc}.

We are now ready to formulate our problem statement:

\begin{definition}{\textbf{Learning problem.}}
{\em Given} a training set $D \subseteq \mathcal{X} \bigtimes \mathcal{Y}$, a set of domain constraints $\mathcal{K}$, a loss function $\mathcal{L}$, and a hypothesis space containing functions $f_\w: \mathcal{X} \rightarrow \mathcal{Y}$; {\em find} $\w$ such that $f_{\w}$ satisfies constraints in $\mathcal{K}$ and $\mathcal{L}(f_\w, D)$ is minimal among all such $f_{\w}$.
\label{def:learning_problem}
\end{definition}
$f_\w$ is assumed to be a feedforward neural network.
The output layer is real-valued without any activation, and a softmax layer is used to calculate class probabilities for classification problems.
The language of the constraints in $\mathcal{K}$ is a subset of first-order logic which allows for universal quantifiers.
In practice, we focus on the constraints of the form in equation \ref{equation:constraint}, and $\mathcal{K}$ can be a set of multiple such universally quantified constraints.

\section{Background - Satisfiability Descent (SaDe)}\label{background:sade}

Satisfiability (SAT) is the problem of finding a solution that satisfies a given Boolean formula, e.g. $\neg a \lor b$.
Satisfiability Modulo Theories (SMT) extend SAT such that formulas can be expressed in other theories, such as Real Arithmetic, e.g., $(a + b > 3)\wedge(b < 1)$ with real $a$ and $b$.
Maximum Satisfiability (MaxSMT) generalizes the SMT problem: given a set of hard constraints $\mathcal{H}$ and soft constraints $\mathcal{S}$, it aims to find a solution that satisfies all constraints in $\mathcal{H}$ and as many as possible in $\mathcal{S}$.

\begin{figure}[t]
\centering
\subfigure[]{\includegraphics[width=.3\textwidth]{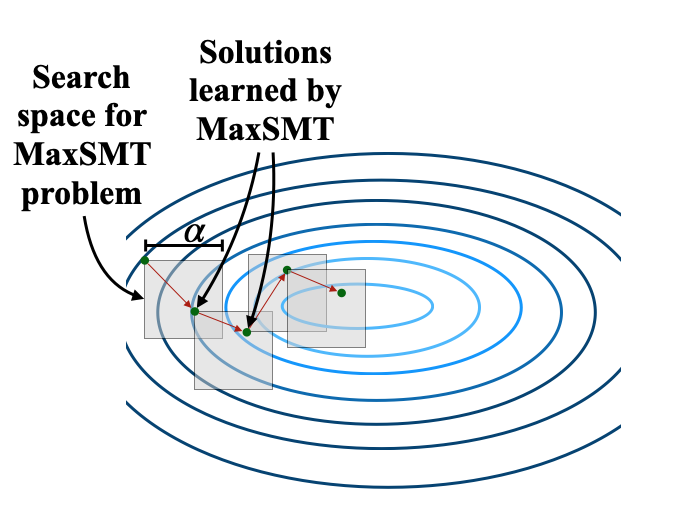}}
\subfigure[]{\includegraphics[width=.34\textwidth]{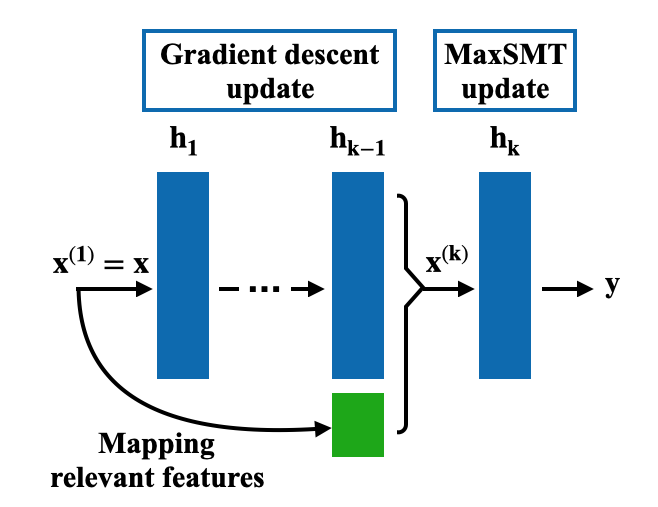}}
\caption{
\textbf{(a.) Illustration of SaDe:} each grey quadrant represents the local search space for the MaxSMT problem ({\em maximal step size: $\alpha$}), defined by the gradients of the loss, the green points are the solutions found with MaxSMT; 
\textbf{(b.) DeepSaDe Architecture:} Last layer is updated using the MaxSMT framework and the layers before are updated with gradient descent. The input features relevant to the domain constraint (in green) are mapped to the penultimate layer via skip connections.
}
\label{fig:sade}
\end{figure}

Our work builds on {\em SaDe} \cite{sade} which is a learning algorithm that can enforce constraints in linear models and guarantee satisfaction.
SaDe modifies the parameter update process of the mini-batch gradient descent \cite{goodfellow2016deep}.
Unlike gradient descent, which updates the solution in the direction that minimizes the loss, SaDe solves a MaxSMT problem to find the solution at each iteration, which is formulated in such a way that its solution satisfies the domain constraint and is close to the solution that gradient descent might lead to.
At each iteration, a local search space is defined for the MaxSMT problem around the previous solution.
This search space is a fixed-sized n-cube where each edge of the n-cube is a hyperparameter called {\em maximal step size} that upper bounds the size of the update in each dimension (illustrated for two parameters in figure \ref{fig:sade}(a)).
SaDe iteratively improves the performance while learning solutions that satisfy the constraint.
Formulations of the MaxSMT problem and the local search space at each iteration are provided next.

{\bf Formulation of MaxSMT problem:}
A MaxSMT problem is defined for a batch of instances in each iteration, where the soft constraints encode a certain quality of fit of the model on the instances in the batch, and the domain constraints are hard constraints.

A soft constraint, for a given instance $(\x, y)$, is defined as a logical constraint that is fulfilled when the prediction of the model $f_\w$ for $\x$ is ``sufficiently consistent'' with the true label $y$.
For regression, given some error $e$, a soft constraint is defined as: $|y - e| \leq f_\w(\x) \leq |y + e|$.
For classification, the sign of $f_\w(\x)$ is assumed to be the indicator of the class and the magnitude indicates the certainty of prediction, the soft constraint takes the form (for a threshold $\tau$):
\begin{eqnarray*}
& &  f_\w(\x) > \text{ } \tau   \qquad \text{if $y = 1$}\\
& &  f_\w(\x) < -\tau   \qquad \text{if $y = -1$}
\end{eqnarray*}
For each instance, multiple soft constraints are formulated for different values of the error $e$ or threshold $\tau$.
Satisfying a maximum of these soft constraints, which is what the MaxSMT problem tries to achieve, correlates with minimizing the prediction loss.
Thus, the solution to this MaxSMT problem at every iteration reduces the loss while satisfying the domain constraints.

{\bf Local search space:}
At every iteration, the MaxSMT problem searches for the next solution in a local search space defined by the n-cube around the previous solution.
It is encoded as an additional hard constraint in the MaxSMT problem, with a ``box'' constraint, which states that the next solution must be inside the axis-parallel box defined by $\hat{\w}$ and $\hat{\w} - \alpha \cdot \mbox{sgn}({\bf g})$, where the sign function is applied component-wise to a vector (a modified sign function is used where sgn$(0)=1$), $\hat{\w}$ is the previous solution, ${\bf g}$ is the gradient of the loss at $\hat{\w}$, and $\alpha$ is the maximal step size.
The box constraint serves two important purposes. Firstly, it provides a general direction in which the loss is minimized, as the box in each dimension is aligned with the negative gradient. Secondly, it stabilizes learning by limiting the size of the updates.
Interestingly, using a regularized loss instead of a standard loss doesn't make a difference, as the constraint remains fulfilled throughout the training steps. With regularization, the regularization factor only becomes active if an instance violates the constraint. However, in SaDe, this never happens during the learning process, making the regularization always $0$.

SaDe is limited to training linear models. 
Training neural networks with the same procedure would require solving the MaxSMT problem with highly non-linear soft constraints, e.g.  $|y - e| \leq f_\w(\x) \leq |y + e|$ where $f_\w(\x)$ is the network, which is not possible with state-of-the-art SMT solvers.

\section{DeepSaDe: Deep Satisfiability Descent}\label{deepsade}

We now present our approach {\em DeepSaDe}, which utilizes the MaxSMT framework proposed in \citet{sade} to train neural networks with constraints.
DeepSaDe exploits the structure of neural networks, which transform the input domain through a series of non-linear layers before a final linear layer maps it to the output. 
As the network output only explicitly depends on the last layer, enforcing the constraint on the last layer is sufficient to enforce the constraint on the network.
DeepSaDe, therefore, uses batch learning with a hybrid procedure that uses MaxSMT to determine the weights in the last layer and gradient descent for all other weights.
In the following, we introduce some notation before we formalize the MaxSMT problem in the last layer, and then detail the learning algorithm. 

$f_\w$ is a fully-connected neural network with $k$ layers such that $f_\w(\x) = h_k(h_{k-1}(...h_1(\x)...))$ for input $\x \in \mathcal{X}$, where $h_n$ is the $n^{th}$ layer.
The input to the $n^{th}$ layer is $\x^{(n)} = h_{n-1}(...h_1(\x)...)$, where $\x^{(1)} = \x$ and the latent input space for the $h_n$ is represented by $\mathcal{X}^{(n)}$, where $\mathcal{X}^{(1)} = \mathcal{X}$.
The size of a layer, $|h_n|$, is the number of neurons in it.
Weight and bias parameters of $h_n$ are denoted by matrices $W^{(n)}$ and $B^{(n)}$ of dimensions $|h_{n-1}| \times |h_n|$ and $1 \times |h_n|$ respectively ($|h_{0}| = |\x^{(1)}|$).
Elements of $W^{(n)}$ and $B^{(n)}$ are referred with lower-case letters.

\subsection{Formulation of the maximum satisfiability problem at the last layer of the network}\label{sec:maxsmt_formulation}
To formalize the MaxSMT problem for the last layer, it is necessary to express both the soft constraints and the hard constraint (domain constraint) in terms of the latent inputs $\x^{(k)}$ rather than the original inputs $\x^{(1)}$. The soft constraint at the last linear layer can be formulated in a similar way to what was presented in section \ref{background:sade}. Specifically, for a regression task, the constraint is given by $|y - e| \leq h_k(\x^{(k)}) \leq |y + e|$, whereas for classification, it takes the form of:
\begin{eqnarray*}
& &  h_k(\x^{(k)}) > \text{ } \tau   \qquad \text{if $y = 1$}\\
& &  h_k(\x^{(k)}) < -\tau   \hspace{14pt} \text{if $y = -1$}
\end{eqnarray*}
To formulate the hard domain constraints, we propose a method for translating each original constraint
K (equation \ref{equation:constraint}) into a constraint K' for the MaxSMT problem at the last layer, given the parameters of layers
$h_1, \ldots, h_{k-1}$, such that a solution to K', combined with $h_1, \ldots, h_{k-1}$, is a solution to K.
\begin{equation}\label{translated-constraint}
\text{K}': \forall \x' \in \bigtimes_{i=1}^{|h_{k-1}|} [l^{(k)}_i, u^{(k)}_i], \x' \models P \Rightarrow  h_k(\x') \models C     
\end{equation}
Here $\x'$ represent the quantified latent variable in the domain of $\mathcal{X}^{(k)}$ bounded by $\bigtimes_{i=1}^{|h_{k-1}|} [l^{(k)}_i, u^{(k)}_i]$. We explain how this translation is achieved in the next two paragraphs. Given a network, with this translation, the last layer of a network can be updated to satisfy the constraint by the network.

\textbf{Domain Bound Propagation:}
To formulate $\text{K}'$, we first consider the bounds of the quantified variable for the latent space of $\mathcal{X}^{(k)}$.
The bounds of the latent space must be such that enforcing the constraint within these bounds enforces the constraint on the original input bounds.
To construct such latent bounds, we rely on interval arithmetic \cite{sunaga1958theory}. This involves calculating the bounds of the output of a layer based on the bounds of the input and the layer parameters, such that any input to the layer takes the output value within the output bounds.
Given the lower and upper bounds $l^{(n)}$ and $u^{(n)}$ for the input $\x^{(n)}$ of the layer $h_n$, following the approach used in \citet{gowal2018effectiveness}, the bounds for the output $\x^{(n+1)}$ are computed as (more details in Appendix A.1):
\begin{equation*}
    \begin{split}
        l^{(n+1)}_i = \text{act}(b^{(n)}_{i} + \sum_{j: w^{(n)}_{j,i} \geq 0} w^{(n)}_{j,i} l^{(n)}_j + \sum_{j: w^{(n)}_{j,i} < 0} w^{(n)}_{j,i} u^{(n)}_j) \\
        u^{(n+1)}_i = \text{act}(b^{(n)}_{i} + \sum_{j: w^{(n)}_{j,i} \geq 0} w^{(n)}_{j,i} u^{(n)}_j + \sum_{j: w^{(n)}_{j,i} < 0} w^{(n)}_{j,i} l^{(n)}_j)
    \end{split}
\end{equation*}
Where `act' is the activation function (like ReLu, Sigmoid, and Tanh).
Given the bounds of the input space, the bounds for the latent space $\mathcal{X}^{(k)}$ can be calculated recursively. 
Enforcing a constraint within the latent bounds enforces the constraint on any input within the input bounds.

\textbf{Identity Mapping of Relevant Features:}
The translation also takes into account that some domain constraints may be dependent on the input space, e.g., ``{\em if an object comes in front of the vehicle, the vehicle must stop}''.
To encode such conditions in $\text{K}'$ (i.e., $\x' \models P$), we make the relevant features, i.e. the features that are needed to encode the property $P$, available at the latent space $\mathcal{X}^{(k)}$.
For this, we use skip-connections \cite{he2016deep} that map these features to the second to last layer $h_{k-1}$ using an identity mapping, as illustrated in figure \ref{fig:sade}(b).
The network, consequently, is no longer fully-connected and these features take the same value as the input.
These mapped features become a part of $\mathcal{X}^{(k)}$ and the input property can be expressed at the last layer.
These features are identified in advance.
Finally, as the output of the network is the same as the output of the last layer, the constraint on the network output can be encoded with $h_k(\x')$.
This completes the formulation of $\text{K}'$.

\begin{algorithm}[!t]
\caption{Deep Satisfiability Descent (DeepSaDe)}\label{sade}
\textbf{input:} training data $D$, validation data $V$, domain constraints $\mathbf{\mathcal{K}}$, batch size $b$, epochs $e$, loss $\mathcal{L}$, maximal step size $\boldsymbol\alpha$, learning rate $\boldsymbol\eta$, line search steps $\mathbf{s}$\\
\begin{algorithmic}[1]
\STATE $f_{\hat{W}}(.)$: Model initialized with a standard approach
\STATE $\mathcal{K'}$ = TranslateDomainConstraint($\mathcal{K}$, $f_{\hat{W}}(.)$)
\STATE Update $\hat{W}$ such that $\hat{W}^{(k)} \models \mathcal{K'}$
\STATE restart = False; $\hat{W}_{out}$ = undefined; Partition $D$ into batches of size $b$
\FOR{each epoch}
    \FOR{each batch $B$ in $D$}
        \STATE ${\mathcal{G}}= \nabla \mathcal{L}(\hat{W})$
        \STATE $\mathcal{K'}$ = TranslateDomainConstraint($\mathcal{K}$, $f_{\hat{W}}(.)$)
        \IF{restart}
            \STATE Randomly flip the sign of each element of $\mathcal{G}^{(k)}$
        \ENDIF\\
        {\color{blue} {\em // try line search, if no solution use MaxSMT}}
        \STATE let $W_i = \hat{W}^{(k)} - \eta\mathcal{G}^{(k)} i/s$, for $i=1, 2, \ldots, s$ 
        \IF{$\{i \, | \, W_i \models \mathcal{K'}\} \not= \emptyset$} 
            \STATE $\hat{W}^{(k)} = \hat{W}_{\max \{i \, | \, W_i \models \mathcal{K'}\}}$ 
        \ELSE 
            \STATE ${\mathcal{G}_{S}} = sgn({\mathcal{G}^{(k)}})$              \STATE $\mathcal{S}$ = SoftConstraints($h_{k-1}(...h_1(B)...)$)
            \STATE $\mathcal{H} = \{W^{(k)} \in \text{Box}(\hat{W}^{(k)}, \hat{W}^{(k)} - \alpha\mathcal{G}_{S})\} \cup \mathbf{\mathcal{K'}}$ \\
            {\color{blue}{\em // assuming MaxSMT returns undefined if unsat}}
            \STATE $\hat{W}^{(k)} = $ MaxSMT$(\mathcal{S}, \mathcal{H})$
        \ENDIF \\
        {\color{blue}{\em // no solution found, restart in another direction}}
        \IF{$\hat{W}^{(k)} \text{ is undefined}$}
            \STATE restart = True 
        \ELSE
            \STATEnonum {\color{blue}{\em // remember best solution found}}\\
            \IF{$eval(f_{\hat{W}}, V) > eval(f_{\hat{W}_{out}}, V)$}
                \STATE $\hat{W}_{out} =  \hat{W}$
            \ENDIF\\
            {\color{blue}{\em // perform gradient descent on earlier layers}}
            \FOR{$n = 1, ..., k-1$}
                \STATE $W^{(n)} = \hat{W}^{(n)} - \eta*\mathcal{G}^{(n)}$ 
            \ENDFOR
            \STATE restart = False
        \ENDIF
    \ENDFOR
\ENDFOR
\STATE \textbf{return} $f_{\hat{W}_{out}}(.)$
\end{algorithmic}
\label{algorithm:deepsade}
\end{algorithm}

\subsection{Learning Algorithm}\label{sec:learning-algorithm}
We now present the algorithm while referring to the pseudo-code in algorithm \ref{algorithm:deepsade}.
DeepSaDe modifies mini-batch gradient descent \cite{goodfellow2016deep}, where forward and backward passes at every iteration (lines 5-7) are kept the same, but the parameter update is split into two parts. 
First, only the last layer's parameters are updated to satisfy the domain constraint using the MaxSMT formulation presented earlier (lines 8-23) (details in next paragraph). 
Second, the earlier layers are updated using gradient descent to optimize predictive loss (lines 27-29).
Importantly, after the latter update, the network does not guarantee constraint satisfaction due to changes in the latent space of the last layer, used to formulate the MaxSMT problem. 
Therefore, the network before the latter update is used for evaluation and a validation set is used to select the best model (lines 24-26).

For updating the last layer, first, a line search is used along the vector that minimizes the prediction loss, and a fixed number of candidate solutions, from furthest to closest, are checked if they satisfy $\mathcal{K'}$ and first one that does is picked (lines 12-14).
If this search does not yield a solution, the MaxSMT problem is formulated based on the inputs to the last layer, and a solution is searched in the local search space around the previous solution defined based on the gradients calculated during the backward pass (lines 16-19) (similar to the approach in SaDe).
Line search is employed first because checking if a point satisfies a constraint is faster than searching for the solution in a domain, finding a solution by merely checking a few candidate points speeds up learning.

Sometimes a solution for the last layer cannot be found because the MaxSMT problem could not be solved within the local search space, possibly due to the gradient pointing to a solution space that violates the constraint.
In such cases, a restart procedure is initiated where the signs of the gradients are randomly flipped (lines 9-11, 21-23) to randomize the direction of the update. 
This may slightly decrease predictive performance, but it effectively restarts the learning process when it gets stuck.

Since our approach is iterative, starting from an initial configuration that satisfies the domain constraint is crucial. Otherwise, we may begin in a solution space far from the constrained space, leading to no updates. We ensure this by first initializing the network using a standard method \cite{he2015delving} (line 1), and then updating the weights of the last layer that satisfy the translated constraint $\mathcal{K'}$ (lines 2-3). It is important to note that we are solving a satisfiability problem here, not the maximum satisfiability problem, as we are not using any soft constraints for this purpose.

\section{Related Work}
A standard approach for enforcing constraints in ML models is {\it regularization}, where a penalty is added to the prediction loss whenever the model makes a prediction that violates the constraint ($(1-\lambda)*\text{loss} + \lambda*\text{regularization}$).
\citet{xu2018semantic} propose a regularization defined on the weighted model count (WMC) \cite{chavira2008probabilistic} of the constraint defined over the network output. \citet{diligenti2017semantic} and \citet{serafini2016logic} propose a fuzzy logic-based regularization for constraints in first-order logic.
There are many other regularization approaches in the literature, e.g. \citet{fischer2019dl2}, \citet{hu2016harnessing}, \citet{stewart2017label}.
Regularization can enforce a variety of constraints, but does not guarantee constraint satisfaction. 
Additionally, high regularization loss with a large value of $\lambda$ may provide stronger constrain satisfaction but impacts the predictive performance negatively.

Some approaches guarantee constraints by construction but are generally limited to enforcing specific types of constraints. For example, monotonic lattices \cite{gupta2016monotonic}, deep lattice networks \cite{you2017deep}, and COMET \cite{sivaraman2020counterexample} are approaches to enforce monotonicity in neural networks; \citet{leino2022self} \& \citet{lin2020art} propose approaches to satisfy some safety specifications.
A more general approach, MultiplexNet, was proposed in \citet{hoernle2022multiplexnet}.
They use a multiplexer layer to satisfy constraints in disjunctive normal form (DNF).
However, DNF representation is limiting as certain constraints have worst-case representations in DNF which may lead to exponentially many terms.
Additionally, constraints conditioned on the input space cannot be enforced.
Another approach, DeepProbLog \cite{manhaeve2018deepproblog}, trains neural networks within the ProbLog framework, where constraints can be enforced with ProbLog.
However, it is limited to modeling discrete variables, and cannot model regression.

Our work relates to combinatorial optimization approaches as we use a MaxSMT-based approach. 
These approaches, however, except \citet{sade}, are limited to discrete models like decision trees \cite{gunluk2021optimal,bertsimas2017optimal,verwer2019learning,demirovic2022murtree} and decision sets \cite{yu2021learning,ignatiev2021scalable}.
Maximum satisfiability, specifically, has also been used in various ML tasks \cite{berg2019applications,cussens2012bayesian,malioutov2018mlic}.
None of these, however, focus on training neural networks.

There are approaches that rely on the idea of bound propagation, also used in our work, to train adversarially robust neural networks \cite{gowal2018effectiveness,zhang2019towards}.
However, the constraints that can be enforced are limited to input-output bounds (if the input is in a given bound, the output should be in a specific bound).
Our approach is more general and can, in theory, handle any constraint that can be written as an SMT formula.

Finally, there are approaches to verify if a network satisfies a constraint \cite{katz2019marabou,wang2018formal,bunel2018unified}. DeepSaDe trains models that do not require verification because constraints are guaranteed by construction.

\section{Experiments}\label{sec:experiment}
We evaluate multiple use cases in various ML tasks with complex domain constraints.
We first outline the research questions, then describe the use cases, our evaluation method, and finally the results.
\begin{enumerate}[label={}, leftmargin=0.3cm, rightmargin=0.5cm, itemsep=-0ex]
    \item \textbf{Q1:} Can existing methods satisfy constraints for all predictions in practice, even if they don't guarantee it?
    \item \textbf{Q2:} How does the predictive performance of DeepSaDe models compare to the baselines?
    \item \textbf{Q3:} Do the DeepSaDe models have a higher training time compared to the baselines?
\end{enumerate}

\subsection{Use Cases}\label{use_cases}
The selection of use cases is done in order to demonstrate the flexible representation of constraints in DeepSaDe. The selected use cases tackle a variety of learning problems and handle different types of constraints. 
We consider constraints in the form presented in equation \ref{equation:constraint}, where $P$ and $C$ are written as SMT formulas over $\{\forall, \exists, \lor, \land, \neg, \geq, \leq, =\}$.
In principle, any constraint that can be written as an SMT formula and can be solved using existing SMT solvers can be enforced using DeepSaDe.
Use cases UC1, UC2 \& UC3 are from \citet{sade}, UC4 is novel, and UC5 is from \citet{xu2018semantic}.

{\bf UC1:} A multi-target regression to predict 5 household expenses using 13 attributes, with 41417 data instances. We enforce two constraints: {\em ``sum of all the expenses must be smaller than the total household income''} and {\em ``going out expense must be smaller than 5\% of the household income''}.

{\bf UC2:} A binary classification problem of predicting if a person should be given a loan or not based on 13 attributes, with 492 data instances.
We enforce the constraint: {\em ``a person with a salary less than 5000\$ and an absent credit history must be denied a loan''}.

{\bf UC3:} A multiclass classification problem to classify a song to one of 5 music genres based on 13 attributes, with 793 data instances. We enforce the constraint: {\em ``a song by `the Beatles' must be classified as either rock or pop''}.

{\bf UC4:} A multi-label classification problem of identifying the labels from a sequence of 4 MNIST images. 
We enforce the constraint: {\em ``the sum of the predicted labels must be greater than $10$''}. 20000 instances are generated by selecting 4 images at random from the MNIST dataset.

{\bf UC5:} A preference learning problem to predict the preference order of different sushi, with a constraint: {\em ``the prediction must have a coherent preference order''}. The preference order of 6 out of 10 sushi is used to predict the preference order of the remaining 4 sushi. The dataset contains 4926 instances. 
The preference ordering over {\em n} items is encoded as a flattened binary matrix $\{X_{ij}\}$ where $X_{ij}$ denotes that item $i$ is at position $j$.
Under this encoding, each instance has 36 features and 16 targets.

The use cases are intentionally designed with some training instances that violate the constraints. While `fixing' or `removing' these violations might be possible in certain cases (like UC2), it is not feasible with complex constraints. For instance, in UC1, assume that there is a training instance where the sum of expenses exceeds household income. Correcting this precisely is problematic because it is unclear how much each expense should be reduced. This issue gets tougher when dealing with multiple constraints. Additionally, real-world scenarios might have violations that are hard to even identify, such as using biased data in fairness-sensitive applications. Thus, we evaluate DeepSaDe and the baselines in cases with such violations. More information about the use cases is provided in Appendix A.2.
Importantly, We can only impose constraints explicitly based on model parameters: in the case of a self-driving car (section \ref{rep:constraints}), identifying pixels representing a moving object is in itself a predictive task beyond our current scope. Thus, such cases are not included and are left for future research.

\begin{table*}[t]
\begin{center}
\begin{tabular}{c|c|c|c|c|c}
\toprule
{\bf UC}  & {\bf Approach} & {\bf Constraint} & {\bf AdI($\mathbf{\delta = 0.1}$)} & {\bf Accuracy/MSE} & {\bf Training Time (sec)}\\
\toprule
\multirow{2}{*}{UC1} & DeepSaDe & $\mathbf{100_{\pm 0}}$ & $\mathbf{0_{\pm 0}}$ & *$38.36_{\pm 4.59}$ & $102341_{\pm 40198}$\\
 & REG & $93.50_{\pm 2.01}$ & $0.97_{\pm 0.003}$ & *$\mathbf{30.50_{\pm 6.49}}$ & $227_{\pm 70}$\\
\midrule
\multirow{3}{*}{UC2} & DeepSaDe & $\mathbf{100_{\pm 0}}$ & $\mathbf{0_{\pm 0}}$ & $80.04_{\pm 4.29}$ & $447_{\pm 105}$\\
 & SL & $\mathbf{100_{\pm 0}}$ & $0.002_{\pm 0.006}$ & $\mathbf{80.17_{\pm 3.88}}$ & $45_{\pm 30}$\\
 & SBR & $\mathbf{100_{\pm 0}}$ & $0.002_{\pm 0.004}$ & $80.04_{\pm 3.95}$ & $45_{\pm 27}$\\
\midrule
\multirow{3}{*}{UC3} & DeepSaDe & $\mathbf{100_{\pm 0}}$ & $\mathbf{0_{\pm 0}}$ & $80.11_{\pm 4.99}$ & $6580_{\pm 1915}$\\
& SL & $99.97_{\pm 0.03}$ & $0.42_{\pm 0.28}$ & $\mathbf{82.53_{\pm 3.58}}$ & $101_{\pm 45}$\\
& SBR & $99.97_{\pm 0.05}$ & $0.29_{\pm 0.10}$ & $76.51_{\pm 7.50}$ & $196_{\pm 68}$\\
\bottomrule
\end{tabular}
\end{center}
\caption{Results: UC1, UC2 \& UC3 (*MSE, lower MSE value is better)}
\label{tab:results123}
\end{table*}
\begin{table*}[ht]
\begin{center}
\begin{tabular}{c|c|c|c|c|c|c}
\toprule
{\bf UC}  & {\bf Approach} & {\bf Constraint} & {\bf Coherent} & {\bf Flattened} & {\bf Jaccard} & {\bf Training Time (sec)}\\
\toprule
\multirow{2}{*}{UC4} & DeepSaDe & $\mathbf{100_{\pm 0}}$ & $6.62_{\pm 1.72}$ & $78.52_{\pm 1.73}$ &  $62.97_{\pm 2.28}$ & $227928_{\pm 30559}$\\
& FFN & $88.00_{\pm 3.26}$ & $\mathbf{23.94_{\pm 4.25}}$ & $\mathbf{85.81_{\pm 1.18}}$ & $\mathbf{71.55_{\pm 2.40}}$ & $3215_{\pm 2663}$\\
\midrule
\multirow{3}{*}{UC5} & DeepSaDe & $\mathbf{100_{\pm 0}}$ & $\mathbf{11.08_{\pm 2.61}}$ & $67.17_{\pm 1.48}$ & $\mathbf{25.94_{\pm 2.89}}$ & $17586_{\pm 5074}$\\
& FFN & $0.04_{\pm 0.15}$ & $0.01_{\pm 0.04}$ & $\mathbf{75.69_{\pm 0.15}}$ & $13.04_{\pm 1.06}$ & $48_{\pm 10}$\\
& SL & $\mathbf{100_{\pm 0}}$ & $4.06_{\pm 3.33}$ & $63.16_{\pm 2.62}$ & $18.08_{\pm 3.33}$ & $298_{\pm 110}$\\
\bottomrule
\end{tabular}
\end{center}
\caption{Results: UC4 \& UC5}
\label{tab:results45}
\end{table*}

\subsection{Evaluation Methodology}\label{evaluation}

{\bf Evaluation:} 
Constraint satisfaction is typically evaluated by the {\em constraint accuracy} metric used in \citet{xu2018semantic} and \citet{fischer2019dl2}, which corresponds to the percent of instances where the constraint is not violated by the prediction.
Such an evaluation, however, is limited to a finite sample of the population. Hence, as a second measure, we also calculate the {\em Adversity Index (AdI)} (\cite{sade}), which is the fraction of instances for which a counter-example to the constraint can be constructed in the neighborhood, defined by an $l_{\infty}$ ball of radius $\delta$ around the instance.
AdI takes a value between 0 and 1, a higher value of AdI implies that the model violates the constraint on more points similar to data instances.
AdI is calculated on full data (training and test) because this provides more instances to evaluate constraint satisfaction; we want the constraints to be satisfied on all data, not only test data.
For DeepSaDe, both constraint accuracy and AdI are zero by construction but we still calculate it as a sanity check.
We use the neural network verification software {\em Marabou} (\cite{katz2019marabou}) to find counter-examples. However, we only compute the AdI for UC1-3 because {\em Marabou} only handles inequality constraints and is not suitable for UC4-5.

For predictive performance, we use MSE for UC1 and accuracy for UC2-3. 
For UC4-5, we use {\em coherent accuracy} which is the fraction of instances for which the model predicts the entire configuration correctly, {\em flattened accuracy} which is the fraction of individually correct binary labels, and the {\em Jaccard accuracy} which is the average Jaccard index for each multi-label prediction compared to true labels. 
Performance is only evaluated for instances where the true label does not violate the constraint as there can be data instances in practice where this happens and a comparison with such instances makes the evaluation biased.

{\bf Baselines:} For UC2 and UC3, we use regularization baselines based on \citet{xu2018semantic} (SL) and \citet{diligenti2017semantic} (SBR).
For UC1, we design a custom regularization loss REG (details in appendix A.2).
For UC4, we could not find any approaches that can enforce such a constraint. Hence, we simply compare it with a feedforward network (FFN).
For UC5, we choose SL and FFN as baselines. 
In \citet{sade}, the authors use a post-processing baseline to enforce constraints at inference time.
This type of baseline is excluded from our evaluation because, as also noted by \citet{sade}, this baseline is only applicable when the constraints are simple and not a generally applicable method.

{\bf Experimental Setup:} For solving the MaxSMT problem, we implement the Fu-Malik algorithm (\cite{fu2006solving}) over the Z3 solver for NRA (Quantified Nonlinear Real Arithmetic) formulas. 
We ran experiments on an Intel(R) Xeon(R) Silver 4214 CPU @ 2.20GHz machine with 125 GB RAM.
For each use case, we run 5 experiments with 5-fold cross-validation, and the data is split 70/20/10 into train/test/validation.
Every feature is scaled in [0, 1], and the radius $\delta=0.1$ is chosen for AdI, which is significantly smaller than the mean $\ell_{\infty}$ distance between two points: this distance for UC1 is $0.75$, for UC2 is $0.97$, and for UC3 is $0.89$.
For regularization, the smallest value of $\lambda$, in $[0, 1]$, that leads to minimum violations on the validation set is selected via cross-validation.
Refer to appendix A.2 for details on the architectures and hyper-parameters.

\subsection{Results (Tables \ref{tab:results123} \& \ref{tab:results45})}\label{eval_results}
{\bf Constraint Satisfaction:} DeepSaDe finds a model that achieves 100\% constraint accuracy for each use case and AdI = 0 for UC1-3.
For UC1, the constraint accuracy for REG is 93.5\%, and counterexamples can be constructed close to 97\% of the instances (AdI = 0.97).
Similar behavior is seen for UC3 for both SL and SBR, with SBR proving to be more effective in enforcing constraints because of lower AdI. 
For UC2, SL and SBR both lead to 100\% constraint accuracy, but counterexamples can still be constructed as AdI $> 0$.
For UC4 and UC5, FFN fails to satisfy constraints on test set.
For UC5, SL can satisfy the constraint but the predictive performance is much worse than DeepSaDe.
Thus, in general, the baselines, in contrast to DeepSaDe, {\em do not} satisfy domain constraint satisfaction, which answers {\bf Q1}.

\begin{figure}[h]
\centering
\includegraphics[width=.23\textwidth]{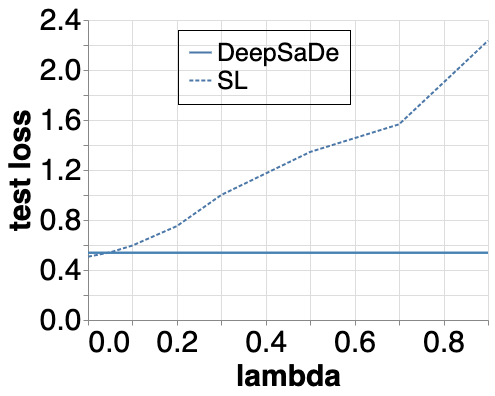}
\caption{Test loss for UC5}
\label{fig:reg-uc5}
\end{figure}

{\bf Predictive Performance:}
DeepSaDe treats the domain constraint as a hard constraint, which limits the solution space to the regions where the constraint is guaranteed.
Thus, the predictive performance of DeepSaDe models can be worse than existing approaches which do not guarantee constraint satisfaction.
For UC1, the predictive performance of DeepSaDe is slightly worse than REG, while the difference is not statistically significant for UC2 \& UC3.
The performance of DeepSaDe is worse for UC4 on all prediction metrics compared to FFN.
For UC5, SL regularization with a high value of $\lambda$ allows for satisfying all the constraints on the test set but performs much worse than DeepSaDe.
To study this further, we plot the prediction loss (cross-entropy) for SL models on the test set for various $\lambda$ between $0$ and $0.9$, averaged over 5 folds, in Figure \ref{fig:reg-uc5}.
For comparison, the average loss for DeepSaDe is also plotted.
DeepSaDe achieves constraint satisfaction in addition to having better performance than SL with a high $\lambda$.
High regularization makes the prediction loss insignificant compared to the regularization loss, leading to worse predictive performance.

In DeepSaDe, at every iteration, the solver tries to satisfy as many soft constraints as possible in addition to satisfying the domain constraints.
DeepSaDe, thus, is a more stable learner compared to regularization with high $\lambda$. 
This answers \textbf{Q2}.
Although the constraint satisfaction with DeepSaDe comes at the cost of predictive performance in some cases, in applications where constraints are crucial, like in safety-critical domains, this may be acceptable.

{\bf Training Time:} 
DeepSaDe requires between 10 to 500 times more time than baselines across use cases.
This is because DeepSaDe solves a MaxSMT problem at each iteration, which makes the training slower than the numerical updates in the baselines.
This positively answers \textbf{Q3}.
In applications where constraints are imperative, training time is less relevant, e.g., a network trained over a week and guarantees safety is still more valuable to the autonomous vehicle compared to one trained for a few hours but cannot do so.
Our work is a starting point of such an approach that combines the exact solving of quantifiers with gradient-based learning.
With further research into solver technology, it can be made more scalable.
Additionally, possible modifications to improve the efficiency of DeepSaDe include using an incomplete MaxSMT approach like stochastic local search \cite{morgado2013iterative} instead of a complete one like Fu-Malik, and using compact latent bounds \cite{wang2018formal}.

\section{Conclusion}
We introduced DeepSaDe, a method for training feedforward neural networks to satisfy a range of constraints through a combination of a satisfiability framework and gradient-based optimization. DeepSaDe effectively addresses various ML tasks and offers adaptable constraint representation, although sometimes with a performance trade-off.
It enforces any constraint in an SMT formula, feasible for the solver. 
While Z3 is our current solver, our framework is adaptable to any MaxSMT solver handling universally quantified constraints.
We believe that evolving solver capabilities will allow DeepSaDe to handle more complex constraints.
Extension of DeepSaDe to other architectures (e.g. CNNs) is left for future work.

\section{Acknowledgement}
This research was jointly funded by the Flemish Government (AI Research Program) and the Research Foundation - Flanders under EOS No. 30992574.

\bibliography{aaai24}
\newpage
\textbf{  }

\newpage

\appendix
\section{Appendix}

\subsection{A.1 \hspace{0.2cm} DeepSaDe: Domain Bound Propagation}
This appendix provides more details on the calculation of the latent bounds in the domain-bound propagation approach in section 4.
The calculation detailed below is based on the interval arithmetic approach (\cite{sunaga1958theory, gowal2018effectiveness}).
Given the lower and upper bounds $l^{(n)}$ and $u^{(n)}$ for the input $\x^{(n)}$ of layer $h_n$, the output of the $i^{th}$ neuron of layer $h_n$ for an arbitrary input $\x^{(n)}$ can be written as (assuming activation):
\[x^{(n+1)}_i = \text{activation}(b^{(n)}_{i} + \sum_{j = 1}^{|h_{n-1}|} w^{(n)}_{j, i}x^{(n)}_{j})\]
Assuming that the activation function is monotonically non-decreasing, which the typically used activations (like ReLu, Sigmoid, Tanh) are, we can bound $x^{(n+1)}_i$ by lower and upper bounds $l^{(n+1)}_i$ and $u^{(n+1)}_i$ where:
\begin{equation*}
    \begin{split}
        l^{(n+1)}_i = \text{activation}(\min_{\x^{(n)}}(b^{(n)}_{i} + \sum_{j = 1}^{|h_{n-1}|} w^{(n)}_{j, i}x^{(n)}_{j})) \\
        u^{(n+1)}_i = \text{activation}(\max_{\x^{(n)}}(b^{(n)}_{i} + \sum_{j = 1}^{|h_{n-1}|} w^{(n)}_{j, i}x^{(n)}_{j}))
    \end{split}
\end{equation*}
In each formula, all terms can be maximized or minimized independently because the domain is an axis-parallel hyperrectangle.
The bounds can be written as:
\begin{equation*}
    \begin{split}
        l^{(n+1)}_i = \text{activation}(b^{(n)}_{i} + \sum_{j = 1}^{|h_{n-1}|} \min_{\x^{(n)}}(w^{(n)}_{j, i}x^{(n)}_{j})) \\
        u^{(n+1)}_i = \text{activation}(b^{(n)}_{i} + \sum_{j = 1}^{|h_{n-1}|} \max_{\x^{(n)}}(w^{(n)}_{j, i}x^{(n)}_{j}))
    \end{split}
\end{equation*}
Note that for any linear function $wx$ with $x \in [l,u]$ is maximal when $x=u$ for positive $w$, and when $x=l$ for negative $w$.
Thus, the bounds can be calculated with the following expressions:

\begin{equation*}
    \begin{split}
        l^{(n+1)}_i = \text{activation}(b^{(n)}_{i} + \sum_{j: w^{(n)}_{j,i} \geq 0} w^{(n)}_{j,i} l^{(n)}_j + \\ \sum_{j: w^{(n)}_{j,i} < 0} w^{(n)}_{j,i} u^{(n)}_j) \\
        u^{(n+1)}_i = \text{activation}(b^{(n)}_{i} + \sum_{j: w^{(n)}_{j,i} \geq 0} w^{(n)}_{j,i} u^{(n)}_j + \\ \sum_{j: w^{(n)}_{j,i} < 0} w^{(n)}_{j,i} l^{(n)}_j)
    \end{split}
\end{equation*}

\begin{figure}[ht]
\centering
\includegraphics[width=.45\textwidth]{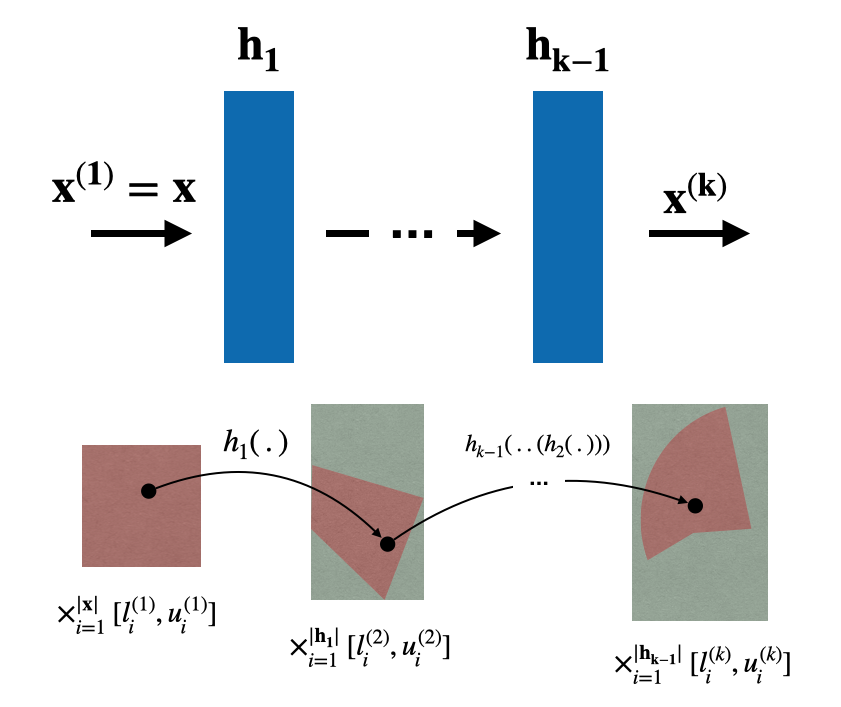}
\caption{Domain Bound Propagation: the bounds for the output of each layer are constructed in such a way that any input to the network takes a value in the latent space within the constructed latent bounds. 
The input domain (in pink), when propagated through the layers, lies completely in the constructed latent domains (in green). 
Thus, enforcing a constraint on the latent bound of the last layer guarantees that the constraint is satisfied on any input within input bounds. In our approach, this construction is essential for achieving constraint satisfaction guarantees for neural networks.}
\label{fig:bound-propagation}
\end{figure}

To conclude, with this approach, the bounds for the output of a layer $h_n$ can be calculated in terms of the bounds of the input and the parameters of the layer.
These bounds can thus be propagated from the input space to the latent space of the last layer.
An illustration of this approach is provided in figure \ref{fig:bound-propagation}.
It is important to note that the bounds calculated with this approach as very wide, but are suitable for our purposes and are cheap to compute.
There are approaches to calculate more compact bounds (\cite{wang2018formal}) that may improve the efficiency of the model, this is left for future work.

\subsection{A.2 \hspace{0.2cm} Experimental Details}
\begin{figure}[t]
\centering
\includegraphics[width=.4\textwidth]{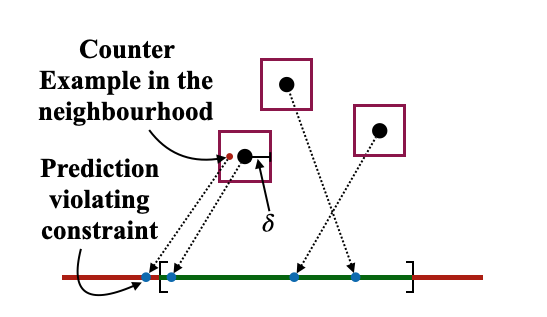}
\caption{\textbf{Illustration of AdI:} 3 instances are mapped to real value predictions using a neural network. The green region is the output region that satisfies the constraint, the red region does not. It is possible, only for the leftmost data point, to construct a new point in the $\ell_{\infty}$ ball such that its prediction violates the constraint, even though the prediction of the original instance does not. Hence, AdI = $1/3$.}
\label{fig:adi}
\end{figure}

In this appendix,  we provide details of the logical formulation of domain constraints for each use case in SMT, along with information regarding the network architecture and the hyper-parameters selected for each use case.
We also provide an illustration of the Adversity Index (AdI) in figure \ref{fig:adi}.
In the following, for a compact representation, we assume $\hat{y}$ represents the real-valued output of the model.
Additionally, to express the component of a variable $\x$, we use subscript with the name of the component (e.g., $\x_{name}$ is the value of the `name' feature of the variable $\x$).
For each network, the ReLU activation is used in the hidden layers. 

{\bf UC1:} 
UC1 is the expense prediction problem with two domain constraints. 
The dataset contains 35317 violations of either of the constraints in 41417 instances.
The `household income' feature is a part of the domain constraints, it is mapped to the last layer with skip connections.
Formulation of the domain constraint at the last layer of the DeepSaDe approach is:
\begin{multline*}
    \mathcal{K'}: \forall \x \in \bigtimes_{i=1}^{|h_{k-1}|} [l^{(k)}_i, u^{(k)}_i], \Big((\text{SUM}(\hat{y}) \leq \x_{household-income}) \\ \land (\hat{y}_{going-out} \leq 0.05*\x_{household-income}) \Big)
\end{multline*}

For the baseline REG, we use two regularization parameters $\lambda$ and $\gamma$ for each constraint with the weight on the predictive loss being $(1 - \lambda - \gamma)$:
\begin{multline*}
    \mathcal{L}= (1 - \lambda - \gamma) * \text{MSE} + \lambda * \frac{1}{\|D\|} * \sum_{\x \in D}\max(0, \sum f_{\w}(\x) \\ - \x_{income}) + \gamma * \frac{1}{\|D\|} * \sum_{\x \in D} \max(0, f_{\w}(\x)_{going\_out} \\ - 0.05*\x_{income})
\end{multline*}
Both $\lambda$ and $\gamma$ are tuned in the range of $[0, 0.49]$ using the 5-fold nested cross validation.
For each model, a network with three hidden layers with sizes 50, 50, \& 14 is considered. 
The learning rate for the baseline and DeepSaDe is selected to be $0.001$ with a batch size of $5$.
For REG, the mode is trained for 10 epochs while for DeepSaDe the model is trained for 5 epochs.
The maximal step size for DeepSaDe is selected to be 0.1 and the error threshold of $[0.1]$ is used for defining the soft constraints.

{\bf UC2:}
UC2 is the loan prediction task that is modeled by a binary classification task. The dataset contains 40 violations of the constraint in 492 instances. Features `income' and `credit history' are mapped to the last layer with identity mapping. 
Formulation of the domain constraint at the last layer of the DeepSaDe approach is:
\begin{multline*}
    \mathcal{K'}: \forall \x \in \bigtimes_{i=1}^{|h_{k-1}|} [l^{(k)}_i, u^{(k)}_i], \Big((\x_{income} < 5000) \land \\ (\x_{credit\_history} = 0)\Big)  \implies \Big(\hat{y}_{no-loan} > \hat{y}_{loan} \Big)
\end{multline*}
For each model, a network with three hidden layers with sizes 50, 30, \& 10 is considered. For SL and SBR, the learning rate of 0.1, and batch size of 5 are selected and the models are trained for 400 epochs.
For DeepSaDe, the learning rate of 0.1, maximal step size of 0.1, batch size of 5, and the margins ($\tau$) of $[0, 1, 2]$ are selected, and the models are trained for 50 epochs.

{\bf UC3:}
UC3 is the music genre prediction task which is modeled by a multiclass classification task. 
The dataset contains 41 violations in 793 instances.
Feature `artist' is mapped to the last layer.
Formulation of the domain constraint at the last layer of the DeepSaDe approach is:
\begin{multline*}
    \mathcal{K'}: \forall \x \in \bigtimes_{i=1}^{|h_{k-1}|} [l^{(k)}_i, u^{(k)}_i], (\x_{artist} = \text{the beatles}) \implies \\ \Big((\hat{y}_{classical} < 0) \land (\hat{y}_{electronic} < 0) \land \\(\hat{y}_{metal} < 0) \land (\hat{y}_{pop} > 0 \lor \hat{y}_{rock} > 0)\Big)
\end{multline*}
For each model, a network with three hidden layers with sizes 50, 30, \& 10 is considered. 
For SL and SBR, the learning rate of 0.1, and batch size of 5 are selected and the models are trained for 300 epochs.
For DeepSaDe, the learning rate of 0.1, maximal step size of 0.1, batch size of 5, and the margins ($\tau$) of $[0, 1, 2]$ are selected, and the models are trained for 50 epochs.

{\bf UC4:}
UC4 is the multi-label prediction task of classifying a sequence of four MNIST images to their respective classes. 
The dataset contains no violations of the domain constraint.
The domain constraint in this problem does not depend on any of the input features.
Formulation of the domain constraint at the last layer of the DeepSaDe approach is:
\begin{multline*}
        \mathcal{K'}: \forall \x \in \bigtimes_{i=1}^{|h_{k-1}|} [l^{(k)}_i, u^{(k)}_i], \\\Big(\text{SUM([IF($\hat{y}_i > 0$, i, 0)for i in range(10)])} > 10 \Big)
\end{multline*}
For each model, a network with three hidden layers with sizes 50, 50, \& 10 is considered. 
For FFN, the learning rate of 0.1, and batch size of 10 are selected and the models are trained for 40 epochs.
For DeepSaDe, the learning rate of 0.1, maximal step size of 0.1, batch size of 10, and the margins ($\tau$) of $[0, 1, 2]$ are selected, and the models are trained for 60 epochs.

\begin{figure*}[t]
\centering
\subfigure[]{\includegraphics[width=.35\textwidth]{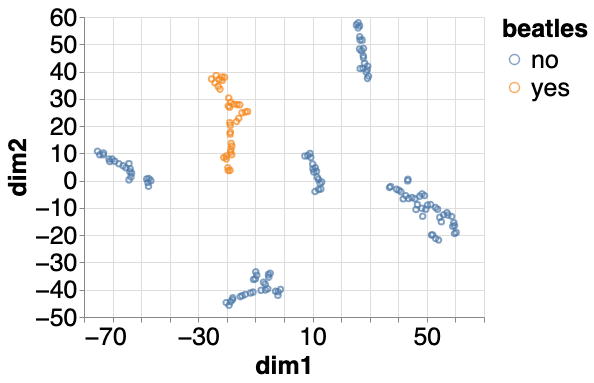}
\includegraphics[width=.375\textwidth]{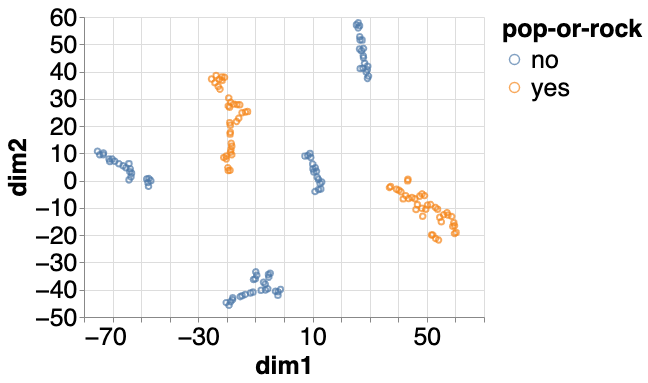}}
\subfigure[]{\includegraphics[width=.35\textwidth]{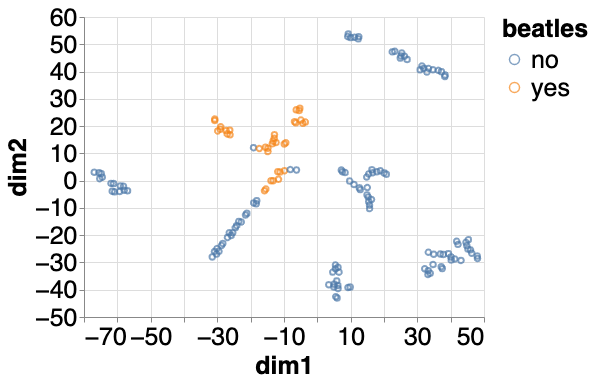}
\includegraphics[width=.375\textwidth]{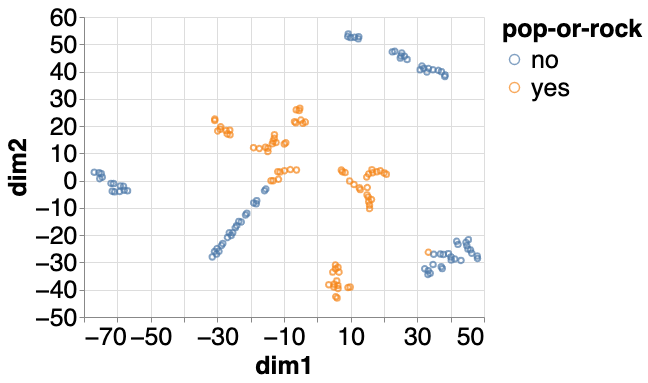}}
\caption{For UC3, a visualization of the latent space of the input to the last layer for the network trained by (a) DeepSaDe and (b) an unconstrained model, on the test data. A 2-dimensional reduction of the latent space using t-SNE is plotted on both plots. In the left plot, the points in orange are the ones that satisfy the input property (i.e., artist = `the Beatles'). In the right plot, the orange points are the ones for which the network predicts the genre as either pop or rock.}
\label{fig:uc3-tsne}
\end{figure*} 

\begin{figure*}[t]
  \centering
  \subfigure[]{\includegraphics[scale=0.2]{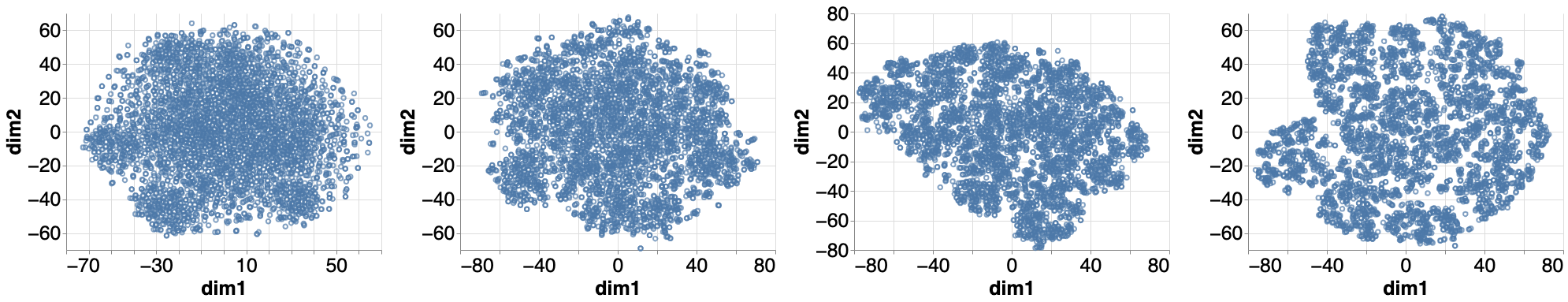}}
  \subfigure[]{\includegraphics[scale=0.2]{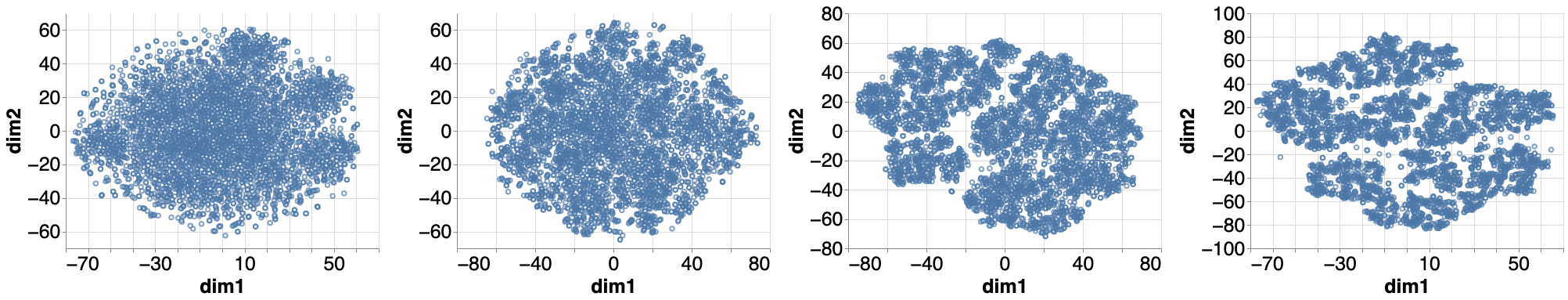}}
  \caption{t-SNE representation of the latent space for (a.) FFN model, and (b.) DeepSaDe model for UC4. From left to right, the input space of the first to the last layer is visualized in a two-dimensional space for a test dataset.
  The latent vectors of the DeepSaDe model exhibit similar behavior as the FFN model, i.e., as the model gets deeper, similar instances can be clustered together by the t-SNE representation.
  This implies that similar to FFN, DeepSaDe is also able to identify patterns in the data while enforcing hard domain constraints which are always satisfied.
  }
  \label{fig:tsne-uc4}
\end{figure*}

{\bf UC5:}
UC5 is the preference learning task with the coherence condition as the domain constraint. 
The dataset contains no violations of the domain constraint.
The domain constraint in this problem does not depend on any of the input features.
The constraints are formulated in DeepSaDe using the following encoding:
\begin{multline*}
    \mathcal{K'}: \forall \x \in \bigtimes_{i=1}^{|h_{k-1}|} [l^{(k)}_i, u^{(k)}_i], \\\Big(\text{PbEq}([\hat{y}_0 > 0, \hat{y}_1 > 0, \hat{y}_2 > 0, \hat{y}_3 > 0], 1) \land \\ \text{PbEq}([\hat{y}_4 > 0, \hat{y}_5 > 0, \hat{y}_6 > 0, \hat{y}_7 > 0], 1)  \land \\ \text{PbEq}([\hat{y}_8 > 0, \hat{y}_9 > 0, \hat{y}_{10} > 0, \hat{y}_{11} > 0], 1) \land \\ \text{PbEq}([\hat{y}_{12} > 0, \hat{y}_{13} > 0, \hat{y}_{14} > 0, \hat{y}_{15} > 0], 1) \land \\ \text{PbEq}([\hat{y}_0 > 0, \hat{y}_4 > 0, \hat{y}_8 > 0, \hat{y}_{12} > 0], 1) \land \\ \text{PbEq}([\hat{y}_1 > 0, \hat{y}_5 > 0, \hat{y}_9 > 0, \hat{y}_{13} > 0], 1) \land \\ \text{PbEq}([\hat{y}_2 > 0, \hat{y}_6 > 0, \hat{y}_{10} > 0, \hat{y}_{14} > 0], 1) \land \\ \text{PbEq}([\hat{y}_3 > 0, \hat{y}_7 > 0, \hat{y}_{11} > 0, \hat{y}_{15} > 0], 1)\Big)
\end{multline*}
Where `PbEq' encodes the exactly-k constraint, $\text{PbEq}([a, b, c], k)$ enforces that exactly $k$ of the boolean constraints $a$, $b$, and $c$ must be true, where $k \in \{1, 2, 3\}$ in this example.
For each model, a network with three hidden layers with sizes 25, 25, \& 10 is considered.
For SL and FFN, the learning rate of 0.1, and batch size of 10 are selected and the models are trained for 40 epochs.
For DeepSaDe, the learning rate of 0.1, maximal step size of 0.1, batch size of 10, and the margins ($\tau$) of $[0, 1, 2]$ are selected, and the models are trained for 40 epochs.

\subsection{A.3 \hspace{0.2cm} Results}
In addition to the experiments presented in the main text, we also perform some supplementary experiments where we visualize the latent space for the networks learned with DeepSaDe.
These experiments demonstrate that DeepSaDe, similar to existing gradient-based approaches, is also able to identify patterns in the latent space, which makes it easier to satisfy the constraint at the last layer.
This experiment also adds validity to DeepSaDe as a learning approach.

{\bf Visualizing the Latent Space:} 
First, we consider the multiclass classification problem of UC3 where the constraint is only enforced for a part of the input space that satisfies the input property (i.e., when the artist = `the Beatles'). In Figure \ref{fig:uc3-tsne}, we plot the t-SNE (t-distributed Stochastic Neighbour Embedding) (\cite{hinton2002stochastic}) transformation of the latent space of the last layer. 
t-SNE captures the structure in the sense that neighboring points in the latent space should be closer in the t-SNE representation. 
In the left figure, data points that satisfy the input property are highlighted (in orange), while in the right figure, data points that are predicted as either `pop' or `rock' by the network are highlighted (in orange). It is shown that the instances that satisfy the input property are clustered together better for DeepSaDe than for the unconstrained model, which implies that enforcing the constraint at the last layer with DeepSaDe enables us to learn the latent space such that enforcing constraints at the last layer becomes easier. Furthermore, the network trained through DeepSaDe satisfies the constraint for each of these instances by predicting the genre to be either pop' or `rock,' as shown in the right figure. However, for an unconstrained network, the model still violates some of the constraints since the constraints are not enforced. We selected UC3 to analyze the distribution of the latent space in the context of the input property, which is difficult to do for use cases UC1, UC4, and UC5, where the constraint is being enforced on all possible predictions.

We perform another analysis for use case UC4 which classifies a collection of MNIST images into respective labels in a multi-label classification task.
Here, the constraint is being enforced for all possible outputs of the network, hence it is difficult to see the impact of constraint satisfaction in the t-SNE representation.
However, in this experiment, we compare how the t-SNE representation of the DeepSaDe model compares with the t-SNE representation for the baseline model (which in this case is a feed-forward neural network without any constraint enforcement). The goal here is to demonstrate the validity of DeepSaDe. 
Figure \ref{fig:tsne-uc4} shows that DeepSaDe exhibits similar behavior in the low dimensional representation of the latent space for each layer when compared with the similar representation of the FFN model. 
DeepSaDe can identify similar points in the latent space (because t-SNE can cluster data points as the model becomes deeper), which means that the network can learn patterns from the data, similar to the FFN while enforcing hard constraints on the predictions which are always satisfied.

\end{document}